\pdfoutput=1

\documentclass[11pt]{article}

\usepackage{emnlp2021}

\usepackage{times}
\usepackage{latexsym}
\usepackage{graphicx} 

\usepackage{booktabs}  
\usepackage[inline]{enumitem}  
\usepackage{multirow}  
\usepackage{comment}

\usepackage{pifont}
\newcommand{\cmark}{\ding{51}}%
\newcommand{\xmark}{\ding{55}}%

\usepackage[T1]{fontenc}

\usepackage[utf8]{inputenc}

\usepackage{microtype}

\usepackage[nameinlink]{cleveref}

\usepackage{url}
\usepackage{xspace}
\newcommand{\safeval}{\textsc{QuestEval}\xspace}

\title{Data-QuestEval: A Reference-less Metric for Data-to-Text Semantic Evaluation}

\author{Clément Rebuffel \thanks{~~Equal contribution}$~~^{1,2}$, Thomas Scialom \footnotemark[1]$~~^{1,3} $, \\
        \textbf{Laure Soulier$^{1}$, Benjamin Piwowarski$^{1}$, Sylvain Lamprier$^{1}$,} \\
        \textbf{Jacopo Staiano$^{3}$, Geoffrey Scoutheeten$^{2}$, Patrick Gallinari$^{1,4}$} \\
        $^1$ Sorbonne Universit\'e, CNRS, LIP6, F-75005 Paris, France \\
        $^2$ BNP Paribas, Paris, France \\
        $^3$ reciTAL, Paris, France \\
        $^4$ Criteo AI Lab, Paris, France
        }

\begin{document}
\maketitle
\begin{abstract}


\safeval is a reference-less metric used in text-to-text tasks, that compares the generated summaries directly to the source text, by automatically asking and answering questions. Its adaptation to Data-to-Text tasks is not straightforward as it requires multimodal Question Generation and Answering systems on the considered tasks, which are seldom available. To this purpose, we propose a method to build synthetic multimodal corpora enabling to train multimodal components for a data-QuestEval metric. 
The resulting metric is reference-less and multimodal; it obtains state-of-the-art correlations with human judgment  on the WebNLG and WikiBio benchmarks. We make data-\safeval's code and models available for reproducibility purpose, as part of the \safeval project.\footnote{\href{https://github.com/ThomasScialom/QuestEval}{https://github.com/ThomasScialom/QuestEval}}

\end{abstract}

\section{Introduction}

Data-to-Text Generation (DTG) aims at generating descriptions in natural language given a structured input, e.g. a table~\citep{Gatt-2018-survey}. Reliability and precision of generated texts is currently regarded as  a  major issue in DTG~\citep{Gardent-2020-book}, with experimental surveys showing that real-life end users of DTG systems care more about accuracy  than about readability~\citep{Reiter-2009-investigation}. 
Neural NLG systems are known to be fluent, but prone to hallucinations \cite{lee2018hallucinations}, i.e. they tend to include nonfactual information.
However, their evaluation remains an open research problem
~\cite{Novikova-2017-need}.

\begin{figure}[t]
\centering
\resizebox{\columnwidth}{!}{
    \includegraphics[width=0.5\textwidth]{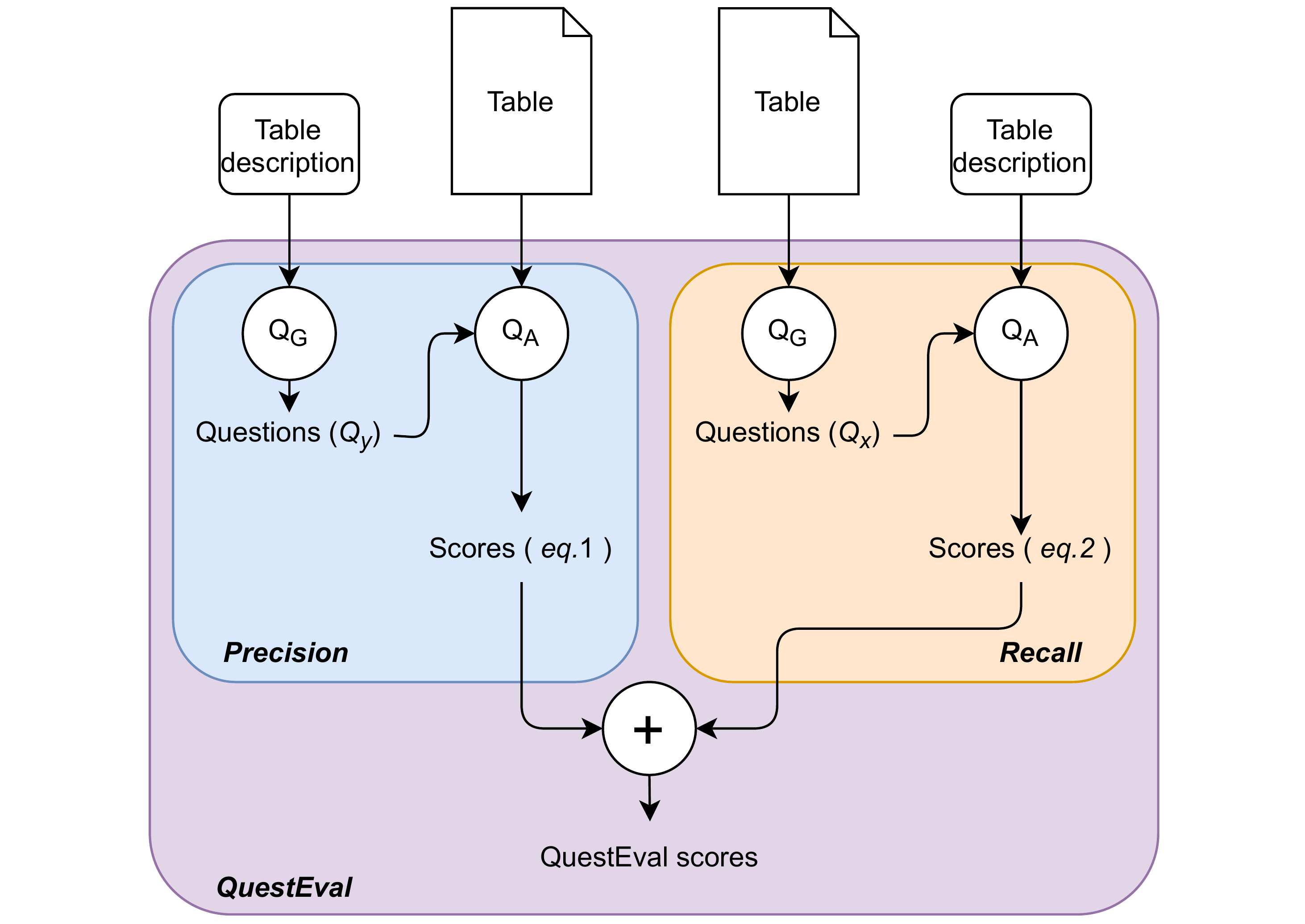}
    }
    \caption{\textbf{Data-\safeval Flowchart.} Figure adapted from the work of \citet{scialom2021safeval} (equation numbers refer to equations in the original paper).}
    \label{fig:questeval-flowchart}
    \vspace{-4mm}
\end{figure}

A recent approach, QuestEval \cite{scialom2021safeval} has shown significant improvement over standard metrics on Summarization tasks. 
To measure semantic matching between an evaluated summary and its source document, QuestEval relies on Question Generation and Answering (QG/QA) systems. 
As illustrated in \Cref{fig:questeval-flowchart}, a QG system generates a set of relevant questions conditioned on the source document, which are then asked on its generated summary. Conversely, questions generated from the summary are answered using only the source input. If the answers provided by the QA systems are correct, the summary is deemed consistent with its source document.

Can QuestEval be adapted for evaluation on DTG tasks? So far, QuestEval's QG/QA systems have been trained on a purely textual dataset, SQuAD \cite{rajpurkar2016squad}, which restricts the evaluation to comparisons between two texts. Unfortunately, DTG inputs are of different modalities than text (e.g. structured tables).
In the absence of specific multimodal-QA datasets, how can one obtain these multimodal-QG/QA models required for a data-QuestEval?  
%

To fill this gap, we propose an effective method for creating synthetic multimodal-QG/QA datasets, by relying only on existing, purely textual, QG/QA datasets. Trained on such synthetic multimodal datasets, QA and QG models can now be used in QuestEval, enabling direct comparison between an evaluated text and its structured input, removing the need for costly gold references. 
Furthermore, this method does not rely on any task-specific annotated QA dataset, which makes the approach general and suitable for any DTG task.

\section{Related Work}

Based on n-grams similarity between an evaluated text and its gold references, BLEU~\cite{Papineni-2002-bleu} and PARENT \citep{Dhingra-2019-parent} are the two standard metrics reported in DTG evaluations. Beyond n-grams, \citet{opitz2020towards} proposed to use a model trained on the reverse task, i.e. text to data reconstruction, and to compare the two data generated i) from the reference, and ii) from the hypothesis. 
\citet{Zhang-2020-bertscore} introduced BERTScore, where texts are compared given the contextual BERT representations of their tokens. 
However, all these metrics require gold references. In a first attempt for a reference-less evaluation metric in DTG, \citet{Dusek-2019-ranting} proposed to train a neural model to directly predict the human ratings; this requires a significant amount of human annotated data, and is eventually biased towards the annotator rather than the task \citep{geva-2019-modeling}. %

Concurrently, a family of reference-less metrics has emerged in Summarization \citep{chen2017semantic, Scialom-2019-answers}. The amount of information shared between two texts is measured by generating questions on the source text, and asking them on the evaluated text. In its recent extension, \safeval \cite{scialom2021safeval} was shown to overperform standard metrics in Text-vs-Text tasks such as the evaluation of summaries.

Unfortunately, text-QG/QA systems are not usable off the shelf for tasks in other modalities which require multimodal-QG/QA systems. Further, there is significant variability in structures (e.g. tables, knowledge graphs, etc.) and domains (e.g. biographies, sports, etc.) across DTG tasks \citep{Gatt-2018-survey}. Therefore, generalizing \safeval to DTG tasks relying solely on existing multimodal-QG/QA datasets is not a promising direction: the effectiveness of data-QG/QA models would be limited to the specific structures of the few existing data-QA datasets -- e.g. the WikiTableQuestions benchmark \citep{Pasupat-2015-wikitq}; moreover, it is unrealistic to annotate QG/QA corpora for each specific modality/domain.

Conversely, our proposed method generalises to any structured data, given only a text-QA dataset.

\begin{table*}[ht]
    \centering \small
    \begin{tabular}{ccccccc}
    \toprule
    \multirow{3}*{Metric} & \multirow{3}*{Reference-less} & \multicolumn{3}{c}{WebNLG} & \multicolumn{2}{c}{WikiBio} \\
    \cmidrule(lr){3-5} \cmidrule(lr){6-7}
    &&  Fluency & Grammar & Semantic &  Fluency & Semantic \\
    \midrule
    \verb|BLEU|   & \xmark & 41.0   & 41.8  & 51.4  & 0.8 & 8.1 \\
    \verb|PARENT| & \xmark & 47.33  & 50.33 & 63.99 & -1.1 & 9.5 \\
    
    \verb|BERTScore| & \xmark & \textbf{58.5}  & \textbf{63.8} & 60.8 & 11.5 & 8.1  \\
    \verb|GPT-2 Perplexity| & \cmark & 49.4  & 56.1 & 48.7 & 7.5 & 7.0  \\
    
    \verb|text-QuestEval|     &  \xmark  &  55.0  &  59.9  &  71.0  & 11.0 & 15.2   \\
    \verb|data-QuestEval|     &  \cmark  &  57.6  &  60.4  &  \textbf{73.5}  & \textbf{13.2} & \textbf{18.2*}  \\
    
    \verb|OOD-data-QuestEval| & \cmark & 57.27  &  61.1  &  71.52 & \textbf{13.2} & 15.8  \\
    \bottomrule
    \end{tabular}

    \caption{Pearson correlation coefficients of considered metrics against human judgements of Fluency, Grammar and Semantic. On WebNLG, all scores have a p-value $p < 0.05$; On WikiBIO, * indicates $p < 0.05$.}
    \label{tab:different-weight-combinations}
\end{table*}

\section{Our approach}
\label{sec:approach}

\begin{figure}[t]
\centering
\resizebox{\columnwidth}{!}{
    \includegraphics[width=0.5\textwidth]{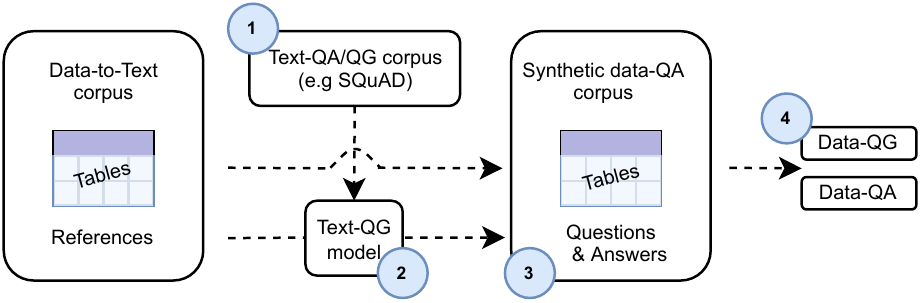}
    }
    \caption{\textbf{Synthetic corpus creation} We are able to create a dataset of (table, question, answer) triples, by transcribing references into questions via a textual QG-model trained on SQuAD. Numerals refer to steps explained in ~\Cref{subsec:refless-safeval}.}
    \label{fig:synth-corpus}
    \vspace{-4mm}
\end{figure}

\subsection{\safeval for Data-To-Text}
\label{subsec:safeval-dtg}

To evaluate semantic matching between two input/output texts (e.g. a document and its summary), \safeval~\citep{scialom2021safeval} proceeds in two steps: 1) a Question Generation system generates a set of questions and (true) answers given the input text; 2) a Question Answering system predicts the (candidate) answers to these questions relying only on the output text currently evaluated. 
Candidate answers are evaluated based on F1-score against the true answers and Semantic Matching is then computed as the mean of all F1-scores. 

To apply \safeval to DTG, and still remain in the textual modality, one can consider a simple baseline: comparing an evaluated description with its (textual) reference, instead of its (data) source. 
Since the predicted description and the reference are both texts, this approach enables us to re-use \safeval in its vanilla form without any multimodal requirements. However, this is not satisfactory, as this metric ignores the structured input, contrary to the original intent of \safeval. Further, this makes the metric dependent on human annotations which may be costly to obtain.  

In the following, we present our proposed method to make \safeval data-compatible, allowing it to measure the similarity directly between a structured input and the evaluated description. 

\subsection{Reference-less multimodal Evaluation} 
\label{subsec:refless-safeval}

To make QG/QA metrics usable for reference-less evaluation on DTG tasks, specific QG/QA datasets are needed to train data-aware systems. 
Relying on an existing corpus is not generalizable: it is unreasonable to expect a multimodal-QA dataset for every DTG dataset. The annotation necessary to build such corpora is costly and time consuming. For this reason, we propose a general approach applicable to any DTG dataset, and requiring no annotated multimodal-QA dataset. The overall process entails four steps (illustrated in \Cref{fig:synth-corpus}): 

\paragraph{Step 1) Textual QG} First, following \safeval, we train a textual QG model on SQuAD.

\paragraph{Step 2) Synthetic Questions} Given the training set of any DTG dataset, composed of (structured-input, textual description) pairs, we generate synthetic questions for each \emph{textual description} using the \emph{textual QG} (from step 1). 

\paragraph{Step 3) Synthetic multimodal-Question dataset} Each example in the training set is constituted of i) the source (i.e. structured data), ii) the textual description, and iii) the synthetic (Question, Answer) pairs generated during step 2. We can therefore match each structured input to its corresponding set of synthetic Questions \& Answers to build a data-QG/QA dataset.

\paragraph{Step 4) Multimodal-QG/QA model training} 
The newly built synthetic multimodal-Question corpus is used to train multimodal QG/QA models. 
For QA, a source corresponds to the structured data and a synthetic question; the target is the corresponding answer. 

QG can be seen as the dual task of QA: any QA dataset can be used as a QG dataset by considering the question as the target. 
To learn representations from structured data, several approaches have been proposed -- e.g. a hierarchical network \cite{fang2019hierarchical}. We adopt the T5 \citep{Kale-2020-t5} paradigm, where any task is considered as a Text-to-Text task: we linearize the tables and encode them directly using T5.

\subsection{Answer Similarity}
\label{subsec:answer_similarity}
In Question Answering, answer correctness (i.e. did the system find the correct answer?) is traditionally measured via F1-score, as popularized by the SQuAD evaluation script~\cite{rajpurkar2016squad}. However, F1-score is based on exact-matching of n-grams and is not accurate when evaluating a correct answer that is realized differently from the reference (e.g. a synonym). This is especially concerning in DTG, where input tables often contain data that are not found verbatim in texts (e.g. ``\textit{Place of birth: France}'' can be realized as ``\textit{She is a French [...]}''). To deal with this issue, we move away from the F1-score and decide to measure answer's correctness via BERTScore~\citep{Zhang-2020-bertscore}, which compares the two contextualized representation of the compared answers. It allows to smooth the similarity function and provides a more accurate view of answer correctness in most DTG settings.

\subsection{Reproducibility}
\label{subsec:reproducibility}
Beyond enabling the comparison of different versions of a given model for a specific project, evaluation metrics make it possible to compare different models altogether between different projects. To avoid inconsistencies in the reporting of \safeval scores, our code follows the guidelines of \cite{post2018call} and produces a short version string that facilitates cross-paper comparisons, identifying the model checkpoints used, preprocessing steps, etc. Reporting this version string will ensure fair comparison across future works.

\section{Experiments}

In this paper, we aim to evaluate the effectiveness of our proposed multimodal adaptation of QuestEval.
Consistently with previous works \citep{Dhingra-2019-parent,Zhang-2020-bertscore}, metric performance is measured by how much it reflects human judgement, assessed via Pearson correlation coefficient.

\begin{table*}[t]
\centering \small
\begin{tabular}{llll}
\toprule
\multicolumn{4}{l}{\textbf{Source:} {[}'101\_helena | discoverer | james\_craig\_watson', 'james\_craig\_watson | deathcause | peritonitis'{]}}                                                                      \\
\multicolumn{4}{l}{\textbf{Reference:} james craig watson , who died from peritonitis , discovered 101 helena .}                                                                                                     \\
\midrule
\midrule
\textbf{Hypothesis}                                                                                                          & \textbf{Generated Questions}              & \textbf{Predicted Answers} & \textbf{Score} \\
\midrule
\begin{tabular}[c]{@{}l@{}}james craero watson is the discoverts of \\ james patson and he died in california .\end{tabular} & Where did james patson die? [...]     & \emph{Unanswerable}*              & 0.0            \\
\midrule
\begin{tabular}[c]{@{}l@{}}james craig watson , who died of \\ peritonitis , discovered 101 helena .\end{tabular}            & Who discovered 101 helena? [...]     & james craig watson        & 1.0 \\
\bottomrule
\end{tabular}
\caption{Example of a structured input and predictions of two different systems, along with automatically generated questions and predicted answers. *the QA system can warn that no answer can be found.}
\label{table:illustration_explainable_questeval}
\vspace{-4mm}
\end{table*}

\subsection{Metrics}

We compare our approach to four automated metrics (two n-gram-based and two neural-based):

\textbf{BLEU~\citep{Papineni-2002-bleu}} compares n-grams between the output and the reference.

\textbf{PARENT~\citep{Dhingra-2019-parent}} is a DTG-specific metric similar to BLEU, that also includes n-grams from the source data, to favour systems that generate true statements, which may not be mentioned in the gold reference.

\textbf{BERTScore~\citep{Zhang-2020-bertscore}} is a neural metric which computes a similarity score for each token in the candidate sentence with each token in the reference sentence, using cosine similarity between the contextualized embeddings.


\textbf{Perplexity} A sentence is scored using the average perplexity of GPT-2~\citep{Radford-2019-gpt2} across all words. 

\textbf{\safeval} 
\verb|text-QuestEval| corresponds to the baseline presented in \Cref{subsec:safeval-dtg}, using textual \safeval to compare evaluated descriptions to their references. In contrast, \verb|data-QuestEval| corresponds to the multimodal version we propose in \Cref{subsec:refless-safeval}, comparing the evaluated description to its structured source. 
For all experiments, unless stated otherwise, we used multimodal-QG/QA models trained on the synthetic corpus corresponding to the evaluation data (e.g. synthetic WebNLG for evaluation on WebNLG). Finally, \verb|OOD-data-QuestEval| was trained on a synthetic E2E dataset \citep{Dusek-2020-e2e} to assess the impact of out-of-domain training.

\subsection{Datasets} 

We evaluate our metric on WebNLG \cite{Shimorina-2018-webnlg} and WikiBio \cite{lebret-etal-2016-neural}. 

\paragraph{The \textbf{WebNLG} data} consists of sets of RDF triples and corresponding descriptions in 16 DBPedia categories (e.g., Airport, or Food). 
Authors of WebNLG provided a set of $2,000$ English descriptions generated by 10 different systems and rated by human annotators on a 3-level Likert scale on three dimensions: \textit{\textbf{Fluency} - does the text sound fluent and natural?}, \textit{\textbf{Grammar} - is the text grammatically correct?}, and \textit{\textbf{Semantic} - does the text correctly represent the meaning in the data?} 

\paragraph{The \textbf{WikiBio} data} consists of biographies paired with the corresponding infoboxes extracted from Wikipedia biography articles. 
WikiBio tables are one order of magnitude larger than tables in WebNLG (see \Cref{tab:dataset-complexities} in supplementary material), and training instances have been built automatically from online sources, resulting in very noisy reference texts \citep{Dhingra-2019-parent,rebuffel2021controlling}.
This increased complexity shines an interesting light on metrics' performances. 
We used the human evaluation from \citet{rebuffel2021controlling}, who collected ratings of three models (one baseline LSTM, and two SOTA systems) on 200 examples of WikiBio, following the protocol of \citet{Shimorina-2018-webnlg} and evaluating two dimensions: Fluency and Semantic.

\subsection{Results and Discussion}
\label{subsec:discussion}

In \Cref{tab:different-weight-combinations}, we report the correlation scores between several metrics and human judgment on the WebNLG and WikiBio benchmarks. 

To the best of our knowledge, this is the first work that evaluates neural metrics (BERTScore, GPT2-Perplexity and \safeval) for DTG. 

\paragraph{WebNLG}
For Fluency and Grammar, neural metrics (i.e. BERTScore, \safeval and Perplexity) dramatically improve the correlations over ngram-based metrics (i.e. BLEU and PARENT).  When comparing BERTScore to \safeval, while the former require a reference as opposed to the latter, the performance is almost comparable. 
For the Semantic aspect, we find that BERTScore correlates less than PARENT, while QuestEval shows a large improvement over the previous best score of PARENT ($73.5$ vs $~64$ resp.).

\paragraph{WikiBIO}
The WikiBIO annotation corpus differs from WebNLG in two ways: i) the input tables are more complex (see \Cref{tab:dataset-complexities} in supplementary material); ii) the systems used for the evaluation of WikiBIO are very recent; their fluency is close to human level -- see Table~4 in \cite{rebuffel2021controlling}. 
This can explain why no metric exhibits a significant correlation for Fluency. 
On Semantic, the only metric that correlates significantly is \safeval, indicating the effectiveness of our proposed method to evaluate DTG. We stress that Semantic is one of the most important dimensions to measure: current models have shown to be fluent but hallucinating \cite{lee2018hallucinations}.

Finally, we observe on both datasets that \safeval performs better using the source than the reference. We hypothesize that some references fail to give accurate descriptions of the tables, which might explain the lower performance (more details in \Cref{appendix:using-ref} of the supplementary materials). 


\paragraph{On the QG/QA potential}
\safeval directly depends on the performances of its QG/QA models.
While not the focus of this study, recent works \cite{chen2020hybridqa, nan2021fetaqa} on multimodal-QA have shown great promise and could lead to further improvement for data-QuestEval. In a larger view, research in Question Answering has been very prolific, and further improvements, either on tables or texts, will undoubtedly lead to improvements in the quality of \safeval's evaluation. We note that in some way, the problematic of evaluating text is removed from the Data-To-Text task, and moved to the QA field, where it is arguably better defined.

\paragraph{On cross-domain adaptation} How would \safeval perform if its multimodal-QG/QA components were to be trained on another synthetic dataset, such as E2E \cite{Dusek-2020-e2e}? In \Cref{tab:different-weight-combinations}, we observe that the results of \verb|OOD-data-QuestEval| on WebNLG are only slightly lower than those of \safeval in-domain, indicating that our approach generalize on similar domains. In contrast, results on WikiBio are not conclusive, highlighting that significant variations on structure and domain leads to dramatic decrease of performances across tasks. 

\paragraph{An interpretable metric}
As noted by \citet{rebuffel2021controlling}, the commonly used metrics compared in this work provide sentence-level information, which can be hard to decipher: for instance, given a $0.5$ PARENT score, which part of the prediction is incorrect?
In contrast, \safeval scores are directly related to its QA's answers for a set of questions generated by its QG. As such, it provides fine-grained, interpretable explanations of its score, via the analysis of the QG's questions and QA's answers (or lack thereof). 
\Cref{table:illustration_explainable_questeval} showcases an example of tabular data with two evaluated descriptions, generated questions, and the predicted answers. 
This emphasizes the explainability of \safeval, an interesting property that we plan to further explore in future work.

\paragraph{On Multilingual extensions}
Experiments of this paper are performed on English corpora. However, in the light of successful QG/QA approaches in different languages, we believe that our work will generalize well to other languages. In particular, languages in which ngram-based metrics perform poorly due to the language's structure (e.g. Korean \citep{lee-etal-2020-reference}) would benefit the most from our approach).

\section{Conclusion}

In this work, we proposed an efficient method to evaluate in a reference-less multimodal setup of Data-to-Text Generation. To train Data-QuestEval on different modalities, we propose a methodology to create synthetic corpora of (data, question, answer) triples from any Data-to-Text Generation task. 
We show that our approach outperforms several existing metrics w.r.t. human judgement on two standard Data-to-Text Generation benchmarks, and performs reasonably well when trained on a different domain. 

While QG/QA modeling per se is out of the scope of this paper, further improvements could be obtained thanks to advances in multimodal QA, a field particularly prolific \cite{chen2020hybridqa,Herzig-2020-tapas, nan2021fetaqa}. In future works, we plan to study how QG/QA systems perform on more complex and abstractive datasets.


\bibliography{references.bib}
\bibliographystyle{acl_natbib}

\clearpage
\appendix

\section{Datasets}
\label{appendix:datasets}

\begin{table}[ht]
    \centering \small
    \scriptsize
    \begin{tabular}{ccccc}
    \toprule
    
     && E2E & WebNLG &  WikiBIO  \\
    \midrule
    \multirow{2}*{table size} &  max  & 8 & 11 & 86 \\
     & mean & 5.37 & 4.5 & 12.42 \\
    \multirow{2}*{target size} & max  & 67 & 445 & 6340 \\
    & mean  & 19.72 & 117.08 & 97.02 \\
    \bottomrule
    \end{tabular}

    \caption{Lengths of inputs and outputs in E2E, WebnNLG and WikiBIO.}
    \label{tab:dataset-complexities}
\end{table}

\section{Implementation Details}
\label{appendix:implementation-details}

For all our experiments, we used SacreBLEU \cite{post2018call} to compute the BLEU score. 
For PARENT \cite{Dhingra-2019-parent}, we used the original implementation that we simply optimized to run on mutli-cpus environement\footnote{\url{https://github.com/KaijuML/parent}}
For BERTScore, we used the original implementation\footnote{\url{https://github.com/Tiiiger/bert_score}}.
The perplexity was computed with the Hugging Face implementation of GPT2-small \cite{wolf2019huggingface}.
We make \safeval available along with the specific DTG models for reproducibility purpose.\footnote{\emph{Anonymous EMNLP submission - hidden URL}}
All the correlations reported in this paper were computed using the SciPy python library \cite{2020SciPy-NMeth}.

\section{A sequence-level evaluation}
\label{appendix:sequence-level}

The DTG community is progressively progressing toward more complex tables, hence longer descriptions. In this context, the community will need metrics able to evaluate long sequences.
As in a DTG task several realizations of a correct description are possible, the number of potentially correct gold-references exponentially raises w.r.t. the length of the sequence. In this context, \safeval becomes particularly interesting, as it loosens the dependence on a specific realization of the source table. As opposed to token-level metrics, \safeval is robust to sentence splitting, word reordering, and now synonyms (see \Cref{subsec:answer_similarity}. Moreover, \safeval is sensible to Fluency given that it uses neural QA/QG systems based on pre-trained Language Models.

\section{QuestEval: using the reference or not?}
\label{appendix:using-ref}

In our ablation studies, \Cref{tab:different-weight-combinations} of the Main Paper, we compare the effect of using or not the gold reference in QuestEval. We can observe that using only the source performs even better than using only the reference. We hypothesise that some references fail to give accurate descriptions of the tables, which might explain the lower performance. This emphasizes the interest for an evaluation metric able to compare the evaluated text directly against the source.  

\end{document}